\documentclass[letterpaper]{article} 
\usepackage{aaai24}  
\usepackage{times}  
\usepackage{helvet}  
\usepackage{courier}  
\usepackage[hyphens]{url}  
\usepackage{graphicx} 
\usepackage{natbib}  
\usepackage{caption} 
\usepackage{algorithm}
\usepackage{algorithmic}

\usepackage{newfloat}
\usepackage{listings}
\DeclareCaptionStyle{ruled}{labelfont=normalfont,labelsep=colon,strut=off} 
\floatstyle{ruled}
\newfloat{listing}{tb}{lst}{}
\floatname{listing}{Listing}
\author{
    Bingjun Luo\textsuperscript{\rm 1},
    Haowen Wang\textsuperscript{\rm 1},
    Jinpeng Wang\textsuperscript{\rm 1},
    Junjie Zhu\textsuperscript{\rm 1},
    Xibin Zhao\textsuperscript{\rm 1},
    Yue Gao\textsuperscript{\rm 1}
}
\affiliations{
    \textsuperscript{\rm 1}BNRist, KLISS, School of Software, Tsinghua University\\
    luobingjun@gmail.com, \{wang-hw21,wjp21,zhujj18\}@mails.tsinghua.edu.cn, \{zxb, gaoyue\}@tsinghua.edu.cn
}

\usepackage{bibentry}

\usepackage{bm}
\usepackage{amssymb}
\usepackage{amsmath}
\usepackage{multirow}
\usepackage{booktabs}
\usepackage{pgfplots}
\pgfplotsset{compat=1.17}

\begin{document}
\title{Hypergraph-Guided Disentangled Spectrum Transformer Networks for Near-Infrared Facial Expression Recognition}
\maketitle

\newcommand{\figleft}{{\em (Left)}}
\newcommand{\figcenter}{{\em (Center)}}
\newcommand{\figright}{{\em (Right)}}
\newcommand{\figtop}{{\em (Top)}}
\newcommand{\figbottom}{{\em (Bottom)}}
\newcommand{\captiona}{{\em (a)}}
\newcommand{\captionb}{{\em (b)}}
\newcommand{\captionc}{{\em (c)}}
\newcommand{\captiond}{{\em (d)}}

\newcommand{\newterm}[1]{{\bf #1}}

\def\figref#1{figure~\ref{#1}}
\def\Figref#1{Figure~\ref{#1}}
\def\twofigref#1#2{figures \ref{#1} and \ref{#2}}
\def\quadfigref#1#2#3#4{figures \ref{#1}, \ref{#2}, \ref{#3} and \ref{#4}}
\def\secref#1{section~\ref{#1}}
\def\Secref#1{Section~\ref{#1}}
\def\twosecrefs#1#2{sections \ref{#1} and \ref{#2}}
\def\secrefs#1#2#3{sections \ref{#1}, \ref{#2} and \ref{#3}}
\def\eqref#1{equation~\ref{#1}}
\def\Eqref#1{Equation~\ref{#1}}
\def\plaineqref#1{\ref{#1}}
\def\chapref#1{chapter~\ref{#1}}
\def\Chapref#1{Chapter~\ref{#1}}
\def\rangechapref#1#2{chapters\ref{#1}--\ref{#2}}
\def\algref#1{algorithm~\ref{#1}}
\def\Algref#1{Algorithm~\ref{#1}}
\def\twoalgref#1#2{algorithms \ref{#1} and \ref{#2}}
\def\Twoalgref#1#2{Algorithms \ref{#1} and \ref{#2}}
\def\partref#1{part~\ref{#1}}
\def\Partref#1{Part~\ref{#1}}
\def\twopartref#1#2{parts \ref{#1} and \ref{#2}}

\def\ceil#1{\lceil #1 \rceil}
\def\floor#1{\lfloor #1 \rfloor}
\def\1{\bm{1}}
\newcommand{\train}{\mathcal{D}}
\newcommand{\valid}{\mathcal{D_{\mathrm{valid}}}}
\newcommand{\test}{\mathcal{D_{\mathrm{test}}}}

\def\eps{{\epsilon}}

\def\reta{{\textnormal{$\eta$}}}
\def\ra{{\textnormal{a}}}
\def\rb{{\textnormal{b}}}
\def\rc{{\textnormal{c}}}
\def\rd{{\textnormal{d}}}
\def\re{{\textnormal{e}}}
\def\rf{{\textnormal{f}}}
\def\rg{{\textnormal{g}}}
\def\rh{{\textnormal{h}}}
\def\ri{{\textnormal{i}}}
\def\rj{{\textnormal{j}}}
\def\rk{{\textnormal{k}}}
\def\rl{{\textnormal{l}}}
\def\rn{{\textnormal{n}}}
\def\ro{{\textnormal{o}}}
\def\rp{{\textnormal{p}}}
\def\rq{{\textnormal{q}}}
\def\rr{{\textnormal{r}}}
\def\rs{{\textnormal{s}}}
\def\rt{{\textnormal{t}}}
\def\ru{{\textnormal{u}}}
\def\rv{{\textnormal{v}}}
\def\rw{{\textnormal{w}}}
\def\rx{{\textnormal{x}}}
\def\ry{{\textnormal{y}}}
\def\rz{{\textnormal{z}}}

\def\rvepsilon{{\mathbf{\epsilon}}}
\def\rvtheta{{\mathbf{\theta}}}
\def\rva{{\mathbf{a}}}
\def\rvb{{\mathbf{b}}}
\def\rvc{{\mathbf{c}}}
\def\rvd{{\mathbf{d}}}
\def\rve{{\mathbf{e}}}
\def\rvf{{\mathbf{f}}}
\def\rvg{{\mathbf{g}}}
\def\rvh{{\mathbf{h}}}
\def\rvu{{\mathbf{i}}}
\def\rvj{{\mathbf{j}}}
\def\rvk{{\mathbf{k}}}
\def\rvl{{\mathbf{l}}}
\def\rvm{{\mathbf{m}}}
\def\rvn{{\mathbf{n}}}
\def\rvo{{\mathbf{o}}}
\def\rvp{{\mathbf{p}}}
\def\rvq{{\mathbf{q}}}
\def\rvr{{\mathbf{r}}}
\def\rvs{{\mathbf{s}}}
\def\rvt{{\mathbf{t}}}
\def\rvu{{\mathbf{u}}}
\def\rvv{{\mathbf{v}}}
\def\rvw{{\mathbf{w}}}
\def\rvx{{\mathbf{x}}}
\def\rvy{{\mathbf{y}}}
\def\rvz{{\mathbf{z}}}

\def\erva{{\textnormal{a}}}
\def\ervb{{\textnormal{b}}}
\def\ervc{{\textnormal{c}}}
\def\ervd{{\textnormal{d}}}
\def\erve{{\textnormal{e}}}
\def\ervf{{\textnormal{f}}}
\def\ervg{{\textnormal{g}}}
\def\ervh{{\textnormal{h}}}
\def\ervi{{\textnormal{i}}}
\def\ervj{{\textnormal{j}}}
\def\ervk{{\textnormal{k}}}
\def\ervl{{\textnormal{l}}}
\def\ervm{{\textnormal{m}}}
\def\ervn{{\textnormal{n}}}
\def\ervo{{\textnormal{o}}}
\def\ervp{{\textnormal{p}}}
\def\ervq{{\textnormal{q}}}
\def\ervr{{\textnormal{r}}}
\def\ervs{{\textnormal{s}}}
\def\ervt{{\textnormal{t}}}
\def\ervu{{\textnormal{u}}}
\def\ervv{{\textnormal{v}}}
\def\ervw{{\textnormal{w}}}
\def\ervx{{\textnormal{x}}}
\def\ervy{{\textnormal{y}}}
\def\ervz{{\textnormal{z}}}

\def\rmA{{\mathbf{A}}}
\def\rmB{{\mathbf{B}}}
\def\rmC{{\mathbf{C}}}
\def\rmD{{\mathbf{D}}}
\def\rmE{{\mathbf{E}}}
\def\rmF{{\mathbf{F}}}
\def\rmG{{\mathbf{G}}}
\def\rmH{{\mathbf{H}}}
\def\rmI{{\mathbf{I}}}
\def\rmJ{{\mathbf{J}}}
\def\rmK{{\mathbf{K}}}
\def\rmL{{\mathbf{L}}}
\def\rmM{{\mathbf{M}}}
\def\rmN{{\mathbf{N}}}
\def\rmO{{\mathbf{O}}}
\def\rmP{{\mathbf{P}}}
\def\rmQ{{\mathbf{Q}}}
\def\rmR{{\mathbf{R}}}
\def\rmS{{\mathbf{S}}}
\def\rmT{{\mathbf{T}}}
\def\rmU{{\mathbf{U}}}
\def\rmV{{\mathbf{V}}}
\def\rmW{{\mathbf{W}}}
\def\rmX{{\mathbf{X}}}
\def\rmY{{\mathbf{Y}}}
\def\rmZ{{\mathbf{Z}}}

\def\ermA{{\textnormal{A}}}
\def\ermB{{\textnormal{B}}}
\def\ermC{{\textnormal{C}}}
\def\ermD{{\textnormal{D}}}
\def\ermE{{\textnormal{E}}}
\def\ermF{{\textnormal{F}}}
\def\ermG{{\textnormal{G}}}
\def\ermH{{\textnormal{H}}}
\def\ermI{{\textnormal{I}}}
\def\ermJ{{\textnormal{J}}}
\def\ermK{{\textnormal{K}}}
\def\ermL{{\textnormal{L}}}
\def\ermM{{\textnormal{M}}}
\def\ermN{{\textnormal{N}}}
\def\ermO{{\textnormal{O}}}
\def\ermP{{\textnormal{P}}}
\def\ermQ{{\textnormal{Q}}}
\def\ermR{{\textnormal{R}}}
\def\ermS{{\textnormal{S}}}
\def\ermT{{\textnormal{T}}}
\def\ermU{{\textnormal{U}}}
\def\ermV{{\textnormal{V}}}
\def\ermW{{\textnormal{W}}}
\def\ermX{{\textnormal{X}}}
\def\ermY{{\textnormal{Y}}}
\def\ermZ{{\textnormal{Z}}}

\def\vzero{{\bm{0}}}
\def\vone{{\bm{1}}}
\def\vmu{{\bm{\mu}}}
\def\vtheta{{\bm{\theta}}}
\def\vphi{{\bm{\phi}}}
\def\vsigma{{\bm{\sigma}}}
\def\va{{\bm{a}}}
\def\vb{{\bm{b}}}
\def\vc{{\bm{c}}}
\def\vd{{\bm{d}}}
\def\ve{{\bm{e}}}
\def\vf{{\bm{f}}}
\def\vg{{\bm{g}}}
\def\vh{{\bm{h}}}
\def\vi{{\bm{i}}}
\def\vj{{\bm{j}}}
\def\vk{{\bm{k}}}
\def\vl{{\bm{l}}}
\def\vm{{\bm{m}}}
\def\vn{{\bm{n}}}
\def\vo{{\bm{o}}}
\def\vp{{\bm{p}}}
\def\vq{{\bm{q}}}
\def\vr{{\bm{r}}}
\def\vs{{\bm{s}}}
\def\vt{{\bm{t}}}
\def\vu{{\bm{u}}}
\def\vv{{\bm{v}}}
\def\vw{{\bm{w}}}
\def\vx{{\bm{x}}}
\def\vy{{\bm{y}}}
\def\vz{{\bm{z}}}

\def\evalpha{{\alpha}}
\def\evbeta{{\beta}}
\def\evepsilon{{\epsilon}}
\def\evlambda{{\lambda}}
\def\evomega{{\omega}}
\def\evmu{{\mu}}
\def\evpsi{{\psi}}
\def\evsigma{{\sigma}}
\def\evtheta{{\theta}}
\def\eva{{a}}
\def\evb{{b}}
\def\evc{{c}}
\def\evd{{d}}
\def\eve{{e}}
\def\evf{{f}}
\def\evg{{g}}
\def\evh{{h}}
\def\evi{{i}}
\def\evj{{j}}
\def\evk{{k}}
\def\evl{{l}}
\def\evm{{m}}
\def\evn{{n}}
\def\evo{{o}}
\def\evp{{p}}
\def\evq{{q}}
\def\evr{{r}}
\def\evs{{s}}
\def\evt{{t}}
\def\evu{{u}}
\def\evv{{v}}
\def\evw{{w}}
\def\evx{{x}}
\def\evy{{y}}
\def\evz{{z}}

\def\mA{{\bm{A}}}
\def\mB{{\bm{B}}}
\def\mC{{\bm{C}}}
\def\mD{{\bm{D}}}
\def\mE{{\bm{E}}}
\def\mF{{\bm{F}}}
\def\mG{{\bm{G}}}
\def\mH{{\bm{H}}}
\def\mI{{\bm{I}}}
\def\mJ{{\bm{J}}}
\def\mK{{\bm{K}}}
\def\mL{{\bm{L}}}
\def\mM{{\bm{M}}}
\def\mN{{\bm{N}}}
\def\mO{{\bm{O}}}
\def\mP{{\bm{P}}}
\def\mQ{{\bm{Q}}}
\def\mR{{\bm{R}}}
\def\mS{{\bm{S}}}
\def\mT{{\bm{T}}}
\def\mU{{\bm{U}}}
\def\mV{{\bm{V}}}
\def\mW{{\bm{W}}}
\def\mX{{\bm{X}}}
\def\mY{{\bm{Y}}}
\def\mZ{{\bm{Z}}}
\def\mBeta{{\bm{\beta}}}
\def\mPhi{{\bm{\Phi}}}
\def\mLambda{{\bm{\Lambda}}}
\def\mSigma{{\bm{\Sigma}}}
\def\mTheta{{\bm{\Theta}}}
\def\mZero{{\bm{0}}}

\newcommand{\tens}[1]{\bm{\mathsfit{#1}}}
\def\tA{{\tens{A}}}
\def\tB{{\tens{B}}}
\def\tC{{\tens{C}}}
\def\tD{{\tens{D}}}
\def\tE{{\tens{E}}}
\def\tF{{\tens{F}}}
\def\tG{{\tens{G}}}
\def\tH{{\tens{H}}}
\def\tI{{\tens{I}}}
\def\tJ{{\tens{J}}}
\def\tK{{\tens{K}}}
\def\tL{{\tens{L}}}
\def\tM{{\tens{M}}}
\def\tN{{\tens{N}}}
\def\tO{{\tens{O}}}
\def\tP{{\tens{P}}}
\def\tQ{{\tens{Q}}}
\def\tR{{\tens{R}}}
\def\tS{{\tens{S}}}
\def\tT{{\tens{T}}}
\def\tU{{\tens{U}}}
\def\tV{{\tens{V}}}
\def\tW{{\tens{W}}}
\def\tX{{\tens{X}}}
\def\tY{{\tens{Y}}}
\def\tZ{{\tens{Z}}}

\def\gA{{\mathcal{A}}}
\def\gB{{\mathcal{B}}}
\def\gC{{\mathcal{C}}}
\def\gD{{\mathcal{D}}}
\def\gE{{\mathcal{E}}}
\def\gF{{\mathcal{F}}}
\def\gG{{\mathcal{G}}}
\def\gH{{\mathcal{H}}}
\def\gI{{\mathcal{I}}}
\def\gJ{{\mathcal{J}}}
\def\gK{{\mathcal{K}}}
\def\gL{{\mathcal{L}}}
\def\gM{{\mathcal{M}}}
\def\gN{{\mathcal{N}}}
\def\gO{{\mathcal{O}}}
\def\gP{{\mathcal{P}}}
\def\gQ{{\mathcal{Q}}}
\def\gR{{\mathcal{R}}}
\def\gS{{\mathcal{S}}}
\def\gT{{\mathcal{T}}}
\def\gU{{\mathcal{U}}}
\def\gV{{\mathcal{V}}}
\def\gW{{\mathcal{W}}}
\def\gX{{\mathcal{X}}}
\def\gY{{\mathcal{Y}}}
\def\gZ{{\mathcal{Z}}}

\def\sA{{\mathbb{A}}}
\def\sB{{\mathbb{B}}}
\def\sC{{\mathbb{C}}}
\def\sD{{\mathbb{D}}}
\def\sF{{\mathbb{F}}}
\def\sG{{\mathbb{G}}}
\def\sH{{\mathbb{H}}}
\def\sI{{\mathbb{I}}}
\def\sJ{{\mathbb{J}}}
\def\sK{{\mathbb{K}}}
\def\sL{{\mathbb{L}}}
\def\sM{{\mathbb{M}}}
\def\sN{{\mathbb{N}}}
\def\sO{{\mathbb{O}}}
\def\sP{{\mathbb{P}}}
\def\sQ{{\mathbb{Q}}}
\def\sR{{\mathbb{R}}}
\def\sS{{\mathbb{S}}}
\def\sT{{\mathbb{T}}}
\def\sU{{\mathbb{U}}}
\def\sV{{\mathbb{V}}}
\def\sW{{\mathbb{W}}}
\def\sX{{\mathbb{X}}}
\def\sY{{\mathbb{Y}}}
\def\sZ{{\mathbb{Z}}}

\def\emLambda{{\Lambda}}
\def\emA{{A}}
\def\emB{{B}}
\def\emC{{C}}
\def\emD{{D}}
\def\emE{{E}}
\def\emF{{F}}
\def\emG{{G}}
\def\emH{{H}}
\def\emI{{I}}
\def\emJ{{J}}
\def\emK{{K}}
\def\emL{{L}}
\def\emM{{M}}
\def\emN{{N}}
\def\emO{{O}}
\def\emP{{P}}
\def\emQ{{Q}}
\def\emR{{R}}
\def\emS{{S}}
\def\emT{{T}}
\def\emU{{U}}
\def\emV{{V}}
\def\emW{{W}}
\def\emX{{X}}
\def\emY{{Y}}
\def\emZ{{Z}}
\def\emSigma{{\Sigma}}

\newcommand{\etens}[1]{\mathsfit{#1}}
\def\etLambda{{\etens{\Lambda}}}
\def\etA{{\etens{A}}}
\def\etB{{\etens{B}}}
\def\etC{{\etens{C}}}
\def\etD{{\etens{D}}}
\def\etE{{\etens{E}}}
\def\etF{{\etens{F}}}
\def\etG{{\etens{G}}}
\def\etH{{\etens{H}}}
\def\etI{{\etens{I}}}
\def\etJ{{\etens{J}}}
\def\etK{{\etens{K}}}
\def\etL{{\etens{L}}}
\def\etM{{\etens{M}}}
\def\etN{{\etens{N}}}
\def\etO{{\etens{O}}}
\def\etP{{\etens{P}}}
\def\etQ{{\etens{Q}}}
\def\etR{{\etens{R}}}
\def\etS{{\etens{S}}}
\def\etT{{\etens{T}}}
\def\etU{{\etens{U}}}
\def\etV{{\etens{V}}}
\def\etW{{\etens{W}}}
\def\etX{{\etens{X}}}
\def\etY{{\etens{Y}}}
\def\etZ{{\etens{Z}}}

\newcommand{\pdata}{p_{\rm{data}}}
\newcommand{\ptrain}{\hat{p}_{\rm{data}}}
\newcommand{\Ptrain}{\hat{P}_{\rm{data}}}
\newcommand{\pmodel}{p_{\rm{model}}}
\newcommand{\Pmodel}{P_{\rm{model}}}
\newcommand{\ptildemodel}{\tilde{p}_{\rm{model}}}
\newcommand{\pencode}{p_{\rm{encoder}}}
\newcommand{\pdecode}{p_{\rm{decoder}}}
\newcommand{\precons}{p_{\rm{reconstruct}}}

\newcommand{\laplace}{\mathrm{Laplace}} 

\newcommand{\E}{\mathbb{E}}
\newcommand{\Ls}{\mathcal{L}}
\newcommand{\R}{\mathbb{R}}
\newcommand{\emp}{\tilde{p}}
\newcommand{\lr}{\alpha}
\newcommand{\reg}{\lambda}
\newcommand{\rect}{\mathrm{rectifier}}
\newcommand{\softmax}{\mathrm{softmax}}
\newcommand{\sigmoid}{\sigma}
\newcommand{\softplus}{\zeta}
\newcommand{\KL}{D_{\mathrm{KL}}}
\newcommand{\Var}{\mathrm{Var}}
\newcommand{\standarderror}{\mathrm{SE}}
\newcommand{\Cov}{\mathrm{Cov}}
\newcommand{\normlzero}{L^0}
\newcommand{\normlone}{L^1}
\newcommand{\normltwo}{L^2}
\newcommand{\normlp}{L^p}
\newcommand{\normmax}{L^\infty}

\newcommand{\parents}{Pa} 

\let\ab\allowbreak

\begin{abstract}
With the strong robusticity on illumination variations, near-infrared (NIR) can be an effective and essential complement to visible (VIS) facial expression recognition in low lighting or complete darkness conditions. However, facial expression recognition (FER) from NIR images presents more challenging problem than traditional FER due to the limitations imposed by the data scale and the difficulty of extracting discriminative features from incomplete visible lighting contents. In this paper, we give the first attempt to deep NIR facial expression recognition and proposed a novel method called near-infrared facial expression transformer (NFER-Former). Specifically, to make full use of the abundant label information in the field of VIS, we introduce a Self-Attention Orthogonal Decomposition mechanism that disentangles the expression information and spectrum information from the input image, so that the expression features can be extracted without the interference of spectrum variation. We also propose a Hypergraph-Guided Feature Embedding method that models some key facial behaviors and learns the structure of the complex correlations between them, thereby alleviating the interference of inter-class similarity. Additionally, we have constructed a large NIR-VIS Facial Expression dataset that includes 360 subjects to better validate the efficiency of NFER-Former. Extensive experiments and ablation studies show that NFER-Former significantly improves the performance of NIR FER and achieves state-of-the-art results on the only two available NIR FER datasets, Oulu-CASIA and Large-HFE.
\end{abstract}

\vspace{-0.3cm}
\section{Introduction}
FER is an important biometric technology in the field of affective computing, which has been widely used in many areas, including health care and social network. The goal of FER is to encode emotional information from facial geometric and texture details. However, under low lighting or complete darkness conditions, VIS light sensors are not suitable for acquiring high-quality images, which inevitably hinders the accuracy of subsequent FER. Fortunately, the NIR spectrum has been proven to be more robust to visible light illumination variations, showing great potential to be an effective complement to VIS FER. Hence, how to perform NIR FER can be a critical problem for real-world systems.

\begin{figure}[t]
  \centering\includegraphics[width=0.99\linewidth]{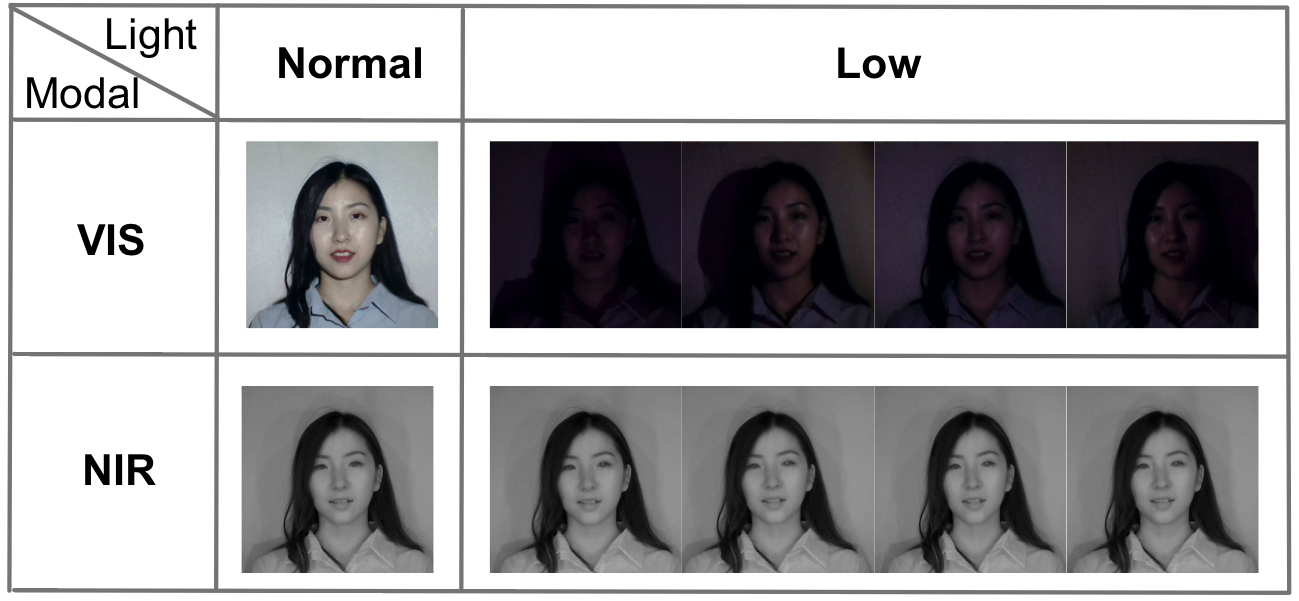}
  \vspace{-2mm}
   \caption{Comparison of VIS and NIR expression images under different lighting conditions. VIS expression images exhibit richer appearance information when captured in normal lighting conditions. However, in the case of low-light conditions, NIR expression images demonstrate superior stability and preservation of visual qualities.}
   \label{fig1}
  \vspace{-4mm}
  \end{figure}
  
As we know, having as many labeled training data as possible is extremely important for designing a deep FER system. Due to the limitation of training samples, NIR FER presents more challenging issues that cannot be solved by simply extending the successful VIS FER method to the NIR spectrum. To acquire sufficient VIS expression images, researchers have made tremendous efforts. From the earliest CK dataset, which collected 486 videos, to Multi-PIE, which collected 755,370 images, both the scale and quality are rapidly improving.  In contrast, up until now, there has been only one available NIR expression dataset, Oulu-CASIA, which contains 480 valid NIR image sequences. This is far from sufficient for obtaining a well-performing NIR FER model.

Since collecting large-scale facial expression images is expensive and time-consuming, starting from scratch may not be feasible. A more ideal solution would be to leverage the rich annotation information from the VIS domain to help alleviate the problem of insufficient data in the NIR domain. To achieve this goal, there are two main challenges as follows: (1) High intra-class variation raised by different geometric and textural details between VIS and NIR images. Since VIS images (380–750 nm) and NIR images (780–1100 nm) are captured through different sensing modalities, even the same person with the same expression in these two modalities will show different patterns. VIS images lack the characteristic spectral information found in NIR images, particularly when it comes to extreme lighting conditions. Additionally, NIR images frequently fail to capture some visual appearances or contents such as the pixels encompassing the cheek and hair. (2) High inter-class similarity. As indicated by Ekman’s Facial Action Coding System (FACS) [43], different expressions often share some key facial behaviors or action units (AU), showing the same visually discernible facial movements. Moreover, individuals in different expression classes also exhibit highly similar facial appearance characteristics.

To address these challenges, we propose a feasible method termed near-infrared facial expression transformer (NFER-Former), which can better extract modality-invariant emotional features from NIR-VIS heterogeneous samples and model high-order correlations among different expression classes for NIR FER.

\textbf{Firstly}, the self-attention orthogonal decomposition (SAOD) module is proposed to decompose the expression and spectrum information from the input image. 
To avoid the limitation of external orthogonality constraint in previous methods, the SAOD module proposes an endogenously orthogonal decomposition strategy based on the linear algebra theorem, and then adapts the Transformer encoder layer to the task of orthogonal subspace decomposition. With the support of Transformer network, the modality-invariant feature is extracted from heterogeneous image samples.

\textbf{Secondly}, the hypergraph-guided feature embedding (HGFE) module is proposed to model the complex inter-class correlations among expression classes. HGFE module explores the AU-expression co-occurrence relationships and represents them as an AU-based knowledge hypergraph. 

With the embedding of hypergraph neural networks (HGNN), knowledge-guided expression features are learned to alleviate the interference of inter-class similarity.

\textbf{Finally}, to further validate the performance of NFER-Former, we collect a new dataset called Large Heterogeneous Facial Expression (Large-HFE). Large-HFE dataset consists of 360 subjects ranging in age from 5 to 70, representing various cultural backgrounds. 

In summary, our main contributions are three-fold:

(1)	We develop a novel method, i.e. NFER-Former, to improve NIR FER performance by leveraging rich annotation information from VIS modality. To the best of our knowledge, this is the first effort to utilize NIR-VIS heterogeneous samples for the improvement of NIR FER performance.

(2)	We collect a new dataset named Large-HFE for better evaluation of NIR FER. The number of subjects in Large-HFE is 4.5 times larger than the previous NIR facial expression dataset. A new benchmark on Large-HFE and Oulu-CASIA is also established to quantitatively evaluate the improvement of NIR FER performance.

(3) We conduct extensive experiments on Oulu-CASIA and Large-HFE. The results demonstrate that the proposed method significantly outperforms state-of-the-art methods.
\vspace{-0.2cm}

\section{Related Work}
\paragraph{Facial Expression Recognition.} With the advancement of deep learning, the accuracy of FER has increased rapidly. Researchers have shifted their focus from designing manual features to improving the FER system's robustness in real-world scenarios. Various challenges, such as occluded facial image \cite{xing2022co}, noisy labels \cite{wang2022ease}, cross-dataset generalization \cite{li2020deeper}, dynamic FER \cite{zhao2021former}, have been addressed through numerous efforts. However, FER in extreme lighting conditions remains an open problem. NIR imaging has shown great potential in outperforming traditional VIS FER in uncontrolled or non-visible light conditions due to its robustness to visible light variations. In this paper, we make the first attempt to utilize NIR-VIS heterogeneous samples to alleviate the overfitting problem on small-scale data, so that better NIR FER performance can be obtained.

\paragraph{Orthognal Subspace Decomposition.} To utilize heterogeneous samples, many efforts have been made to reduce the domain discrepancy between NIR and VIS images. Among them, one mainstream solution is to extract modality-invariant features by mapping the input into a common subspace, which was widely used in heterogeneous face recognition \cite{he2017learning,he2018wasserstein,hu2019disentangled,hu2022domain}. Recently, deep networks are usually proposed to map both NIR and VIS images to a compact Euclidean space and further divide the feature space into two orthogonal subspaces of modality-invariant and modality-specific information respectively by adding additional orthogonality constraints. However, it is hard for such soft orthogonality constraints to enforce orthonormality in models \cite{li2020efficient}. The extra regularization also bring challenges to model optimization \cite{huang2022orthogonal}. In this paper, we propose a novel endogenously orthogonal decomposition strategy to disentangle the modality-invariant and modality-specific information in the feature space, which is more effective and friendly for training.
\vspace{-0.2cm}
\section{Problem Formulation}

Given a NIR-VIS heterogeneous facial expression recognition dataset $\mathcal{D} = \{(\vx^{(i)}, l^{(i)}, y^{(i)})\}_{i=1}^{n}$, where $\vx^{(i)}$ is the image data, $l^{(i)}\in \{N, V\}$ is the modality label (NIR or VIS), $y^{(i)}$ is the corresponding expression label, and $n$ is the number of training samples. 
From the given dataset $\mathcal{D}$, The goal of NIR facial expression recognition is to learn a feature extractor $f(\cdot, \Theta_f)$ that encodes modality-invariant features from both NIR and VIS facial images, and a classifier $g(\cdot, \Theta_g)$ that predicts the expression label from the encoded feature, where $\Theta_f$ and $\Theta_g$ are the parameters of the feature extractor and classifier respectively. During the test process, the recognition performance of the learned feature extractor and classifier is expected to be improved with the joint training of both NIR and VIS facial expression samples.

\vspace{-0.2cm}
\section{Our Model}
\begin{figure*}[htbp]
    \centering
    \vspace{-0.4cm}
    \includegraphics[width=0.99\textwidth]{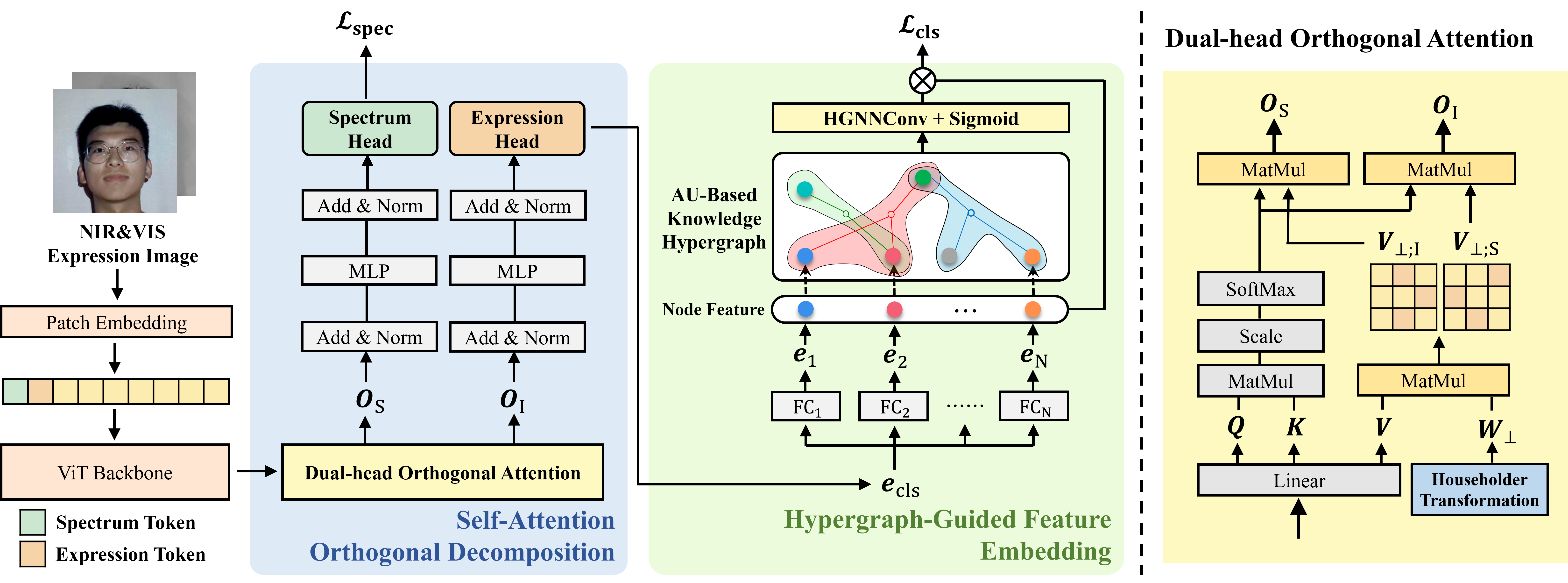} 
    \caption{The main framework of the proposed NFER-Former network for near-infrared facial expression recognition. The framework consists of two modules, including a self-attention orthogonal decomposition module (the blue part) and a hypergraph-guided feature embedding module (the green part). $\mO_S$ and $\mO_I$ denote the modality-specific output and modality-invariant output of the dual-head orthogonal attention block defined in Eq. (\ref{eq:output}).}
    \label{fig:framework}
    \vspace{-0.5cm}
\end{figure*}

To resolve the intra-class variation and inter-class similarity challenge in the NIR-VIS FER task, we propose a transformer-based network named NFER-Former, including Self-Attention Orthogonal Decomposition (SAOD) and Hypergraph-Guided Feature Embedding (HGFE). In this section, we present the overview of our model and the details of the proposed SAOD and HGFE modules.
\vspace{-0.2cm}
\subsection{Model Overview}
As shown in Fig. \ref{fig:framework}, the proposed NFER-Former consists of two components: Self-Attention Orthogonal Decomposition (SAOD) and Hypergraph-Guided Feature Embedding (HGFE).
\textbf{Firstly}, a self-attention orthogonal decomposition mechanism is proposed to remove the spectrum information from the heterogeneous facial expression samples by decomposing the information into orthognoal subspaces. In the SAOD module, the dual-head orthogonal attention is constructed to guarantee the orthogonality of the decomposed features via linear algebra theorem.
\textbf{Secondly}, a hypergraph-guided feature embedding method is proposed to alleviate the interference of inter-class similarity by modeling the complex inter-class correlations using hypergraph. In the HGFE module, the AU-based knowledge hypergraph is constructed to explore the AU-expression co-occurrence relationships and then fed into hypergraph neural networks (HGNN) to learn the knowledge-guided expression features. The whole framework is trained in an end-to-end manner.
\vspace{-0.2cm}
\subsection{Self-Attention Orthogonal Decomposition}\label{sec:SAOD}

In cross-modality FER task, the appearance of facial expressions may be affected by various factors, such as spectrum, identity, pose, etc., which brings large intra-class variations. Similar to the challenge in heterogeneous face recognition \cite{hu2019disentangled}, the spectrum information usually takes up the main part among these factors.

Therefore, the key challenge is to disentangle the expression information and spectrum information from the input image, so that the expression features can be extracted without the interference of spectrum variation.

According to Related Work, a mainstream solution is to decompose the feature map of a backbone network into two orthogonal subspaces respectively, one of which is the modality-specific subspace and the other is the modality-invariant subspace.

However, the previous orthogonal decomposition methods relies on external orthogonality constraints, which are inefficient and unfriendly to the optimization of deep neural networks.

Furthermore, the previous methods are either based on or heavily relied on the property of CNN backbone. 
Therefore, it is still an open problem to incorporate simple yet effective orthogonal decomposition modules with Vision Transformer network for heterogeneous FER.

To solve the above problem, we propose a self-attention orthogonal decomposition (SAOD) module based on the encoder block of Vision Transformer \cite{vaswani2017attention, dosovitskiy2020image}.

In order to conduct effective orthogonal decomposition in the self-attention mechanism, we construct an endogenously orthogonal matrix via linear algebra theorem to parameterize the projection matrix, and further develop a dual-head structure to decompose the input feature into two orthogonal subspaces without imposing any extra constraints. 

Besides, in the input embedding process we append an additional $[spectrum]$ token, which is utilized to indicate the spectrum information of the input image and perform the spectrum classification task. The details are presented as follows.

In each transformer encoder block, the input feature map $\mZ\in\mathbb{R}^{N\times d}$ is linearly transformed to queries, keys, and values $\mQ, \mK, \mV \in \mathbb{R}^{N\times d}$.

Given the matrices of query, key, and value, the attention output matrix $\mO$ is then computed as:
\begin{equation}\label{eq:self-attention}
     \mO = \mbox{softmax}(\frac{\mQ\mK^\top}{\sqrt{d}})\mV
\end{equation}
Inspired by the multi-head mechanism \cite{vaswani2017attention}, a dual-head mechanism is employed in the encoder block. The value matrices $\mV$ are split into two parallel heads as $\mV = [\mV_S, \mV_I]$ where $\mV_S, \mV_I \in \mathbb{R}^{N\times\frac{d}{2}}$ are the modality-specific value matrix and modality-invariant value matrix respectively.

In this way, the input feature $\mZ$ can be decomposed into two representation subspaces using $\mV_S$ and $\mV_I$ respectively according to Eq. (\ref{eq:self-attention}). However, these subspaces are not orthogonal to each other, since $\mV_S$ and $\mV_I$ can be optimized as arbitrary matrices. 

In order to guarantee the orthogonality of the decomposed subspaces, we further introduce an orthogonal projection $\mW_\perp\in\mathbb{R}^{d\times d}$ on the value matrices $\mV_S, \mV_I$ using Householder transformation as
\begin{equation}
     \mW_\perp=\prod_{i=1}^N(\mI-2\frac{\vv_i\vv_i^T}{\Vert \vv_i\Vert^2})
\end{equation}
where $\vv_i\in\mathbb{R}^N (i=1,\cdots,d)$ are learnable parameters. Denote $\mW_\perp=(\vw_1, \cdots, \vw_d)^\top$. By the projection of $\mW_\perp$, $\mV_S$ and $\mV_O$ are transformed as
\begin{equation}
     \mV_{\perp; S} = \mV_S (\vw_1,\cdots, \vw_{d/2})^\top
\end{equation}
\begin{equation}
     \mV_{\perp; I} = \mV_I (\vw_{d/2+1},\cdots, \vw_{d})^\top
\end{equation}
Given the transformed value matrices, we can derive the modality-specific output $\mO_S$ and modality-invariant output $\mO_I$ as
\begin{equation}
     \mO_i = \mbox{softmax}(\frac{\mQ\mK^\top}{\sqrt{d}})\mV_{\perp;i},\ i\in\{S, I\}
     \label{eq:output}
\end{equation}

We will prove the orthogonality of these two decomposed output subspaces in the following.

Firstly, we formulate the proof goal as $\mO_S\mO_I^\top=\mZero$. Denote $\mO_S=(\vo_{S;1}, \cdots, \vo_{S;d})^\top$, $\mO_I=(\vo_{I;1}, \cdots, \vo_{I;d})^\top$. For any $i,j=1, \cdots, d$, row vectors $\vo_{S;i}, \vo_{I;j}$ are any representations in the modality-specific and modality-invariant output subspaces respectively. Therefore, the orthogonality of the two subspaces can be proved by showing that $\vo_{S;i}^\top\vo_{I;j}=0$, which is equivalent to $\mO_S\mO_I^\top=0$.

Secondly, we start from the property of $\mW_\perp$ and derive the proof goal $\mO_S\mO_I^\top=\mZero$. It has been proved that $\mW_\perp$ is endogenously orthogonal and can be optimized as arbitrary orthogonal matrices \cite{huang2022orthogonal}. Therefore, the row vectors of $\mW_\perp$ are orthogonal to each other, i.e., $\vw_i^\top\vw_j=0\ (i\ne j)$. 
By multiplying $\mV_{\perp;S}$ and the transpose of $\mV_{\perp;I}$, we have
\begin{equation}
\begin{aligned}
     \mV_{\perp;S}^\top\mV_{\perp;I} & = \mV_S(\vw_1,\cdots, \vw_{d/2})^\top(\vw_{d/2+1},\cdots, \vw_{d})\mV_I^\top \\
     & = \mV_S
     \begin{pmatrix}
     \vw_1^\top\vw_{d/2+1} & \cdots & \vw_1^\top\vw_{d} \\
     \vdots & \ddots & \vdots \\
     \vw_{d/2}^\top\vw_{d/2+1} & \cdots & \vw_{d/2}^\top\vw_{d}
     \end{pmatrix}
     \mV_I^\top \\
     &=\mZero
\end{aligned}
\end{equation}
Hence, $\mO_S\mO_I^\top=\mZero$.

From the above, it is proved that the input feature map $\mZ$ is decomposed into two orthogonal subspaces $\mO_S$ and $\mO_I$ through the proposed SAOD module. Since the orthogonality is guaranteed endogenously without any extra constraints, the proposed module is more flexible and friendly to the optimization of deep neural networks compared with the existing regularization-based decomposition method \cite{huang2022orthogonal}. Through the great representation power of the self-attention mechanism, orthogonal feature maps are extracted from the input feature map and then fed into the following layers for further processing.

For better disentanglement of expression and spectrum information, we further introduce the spectrum labels to supervise the model besides the original expression labels. Specifically, in the patch embedding process \cite{dosovitskiy2020image}, we append a spectrum token $[spectrum]$ to the existing $[class]$ token. Given the modality-specific output $\mO_S$, a spectrum head implemented by linear and sigmoid layers is proposed to transform the representation of $[spectrum]$ token into the spectrum classification score $\hat{l}$. The spectrum classification loss is defined as the binary cross-entropy loss between the predicted spectrum score $\hat{l}$ and the ground-truth spectrum label $l$ as $\gL_\text{spectrum}$.

\vspace{-0.2cm}
\subsection{Hypergraph-Guided Feature Embedding}\label{sec:HGFE}
After extracting the modality-invariant feature $\mO_I$ by the SAOD module, the framework still faces a great challenge of large inter-class similarity, which lies in the fact that different expressions often share some key facial behaviors \cite{xue2021transfer}. For example, the expressions \textit{happy}, \textit{fearful} and \textit{surprised} usually share the same behavior of parted lips in the mouth, while \textit{sad} and \textit{fearful} share the same behavior of lowered brows in the eyes \cite{du2014compound}. 
The sharing of key behaviors forms complex correlations among different expressions and makes it difficult for the framework to learn discriminative features.

Therefore, it is necessary to model these key facial behaviors and learn the structure of the complex correlations to alleviate the interference of inter-class similarity.

Fortunately, facial action units (AU) can be utilized to represent the key facial behaviors and further construct the complex correlation structure among different expressions. According to the facial action coding system \cite{ekman1978facial}, AU is defined to encode the subtle facial muscle representations and movements, which are widely utilized in the FER field and believed to have strong correlations with facial expression analysis \cite{lien1998automated, tian2001recognizing}. 

Therefore, we propose a hypergraph-guided feature embedding (HGFE) module to address the above challenge of inter-class similarity. The details are described as follows.

Firstly, we introduce the preliminary of hypergraph.
Hypergraph is a generalization of graph where one hyperedge can connect two or more vertices. A hypergraph is defined as $\mathcal{G}=(\mathcal{V}, \mathcal{E})$, where $\mathcal{V}$ is the set of vertices and $\mathcal{E}$ is the set of hyperedges. The hypergraph $\gG$ can be also represented as an incidence matrix $\mH\in\{0,1\}^{|\gV|\times|\gE|}$, with each entry $h(e,v)$ as the indicator of whether hyperedge $v$ contains vertex $c$. If the vertex $v$ is connected by the hyperedge $e$, then $h(v, e)=1$, otherwise $h(v, e)=0$.

Secondly, we formally construct the AU-based knowledge hypergraph. Hypergraph explicitly models the high-order correlations among vertices and edges rather than pair-wise relationships, which is suitable for the representation of complex AU-expression relationships. Therefore, we construct an AU-based knowledge hypergraph to represent the co-occurrence relationships between expressions and prototypical AUs according to \cite{du2014compound}. Denote $v_1, v_2, \cdots, v_N$ as the $N$ prototypical AUs involved in the expression dataset, and $e_1, e_2, \cdots, e_M$ as the $M$ expression classes ($M=6$ for six basic expressions). The AU-based knowledge hypergraph $\gG_k$ is constructed as $\gG_k=(\gV, \gE)$, where $\gV=\{v_1, v_2, \cdots, v_N\}$ and $\gE=\{e_1, e_2, \cdots, e_M\}$. The entry $h(v, e)$ of the incidence matrix $\mH$ is defined as
\begin{equation}
     h(v, e)=\left\{
     \begin{array}{ll}
     1, & \text{if } v \text{ is consistently active in } e \\
     0, & \text{otherwise}
     \end{array}
     \right.
\end{equation}

Thirdly, we decompose the modality-invariant feature and feed it into HGNN \cite{feng2019hypergraph} convolutional layers with the guidance of the constructed knowledge hypergraph. The $[class]$ embedding $\ve_\text{cls}\in \mathbb{R}^d$ is obtained from the modality-invariant feature $\mO_I$ of the SAOD module and then mapped into $N$ branches with fully connected layers as
\begin{equation}
     \ve_i=\text{FC}_i(\ve_\text{cls}), i=1,2,\cdots,N
\end{equation}
where $\ve_i\in \mathbb{R}^{d^{(0)}}$ can be regraded as the vertex feature of $v_i$ in the knowledge hypergraph $\gG_k$. The vertex features are concatenated to form the vertex feature matrix $\mE^{(0)}=(\ve_1, \cdots, \ve_N)^\top\in\mathbb{N}^{N\times d^{(0)}}$ at the input layer. Given the vertex matrix $\mE^{(l)}$ at layer $l$ and the hypergraph $\gG_k$, the result $\mE^{(l+1)}$ at HGNN layer $l$ is calculated as
\begin{equation}
     \begin{aligned}
          \mE^{(l+1)}&=\text{HGNNConv}(\mE^{(l)}, \mTheta^{(l)})\\
          &=
          \text{ReLU}(\mD_v^{-\frac{1}{2}}\mH\mD_e^{-1}\mH^\top\mD_v^{-\frac{1}{2}}\mE^{(l)}\mTheta^{(l)})
     \end{aligned}
\end{equation}
where $\mTheta^{(l)}$ is the learnable parameter matrix at layer $l$, $\mH$ is the incidence matrix of $\gG_k$, $\mD_v\in\mathbb{N}^{N\times N}$ and $\mD_e\in\mathbb{N}^{M\times M}$ are the diagonal matrices of vertex and hyperedge degrees of $\gG_k$ respectively.

Finally, we obtain the output of the HGNNConv layer sequences and utilize it to aggregate the multiple branches for final feature embedding. Through the HGNNConv layer sequences of length $L$, the dimension of the vertex feature is gradually reduced to $1$. The output $\mE^{(L)}\in\mathbb{R}^{N\times 1}$ of the last HGNNConv layer is fed into a sigmoid activation layer to produce the vertex-wise attention weight.

Based on this attention weight, the decomposed branches $\ve_i, \cdots, \ve_N$ are further aggregated in the form of weighted sum to produce the final feature embedding 
\begin{equation}
     \mE_\text{agg}=\sum_{i=1}^N\text{Sigmoid}(\mE_i^{(L)})\cdot\ve_i
\end{equation}

Based on the feature embedding $\mE_\text{agg}$, an expression classification head implemented by single fully connected layer is attached to obtain the final expression prediction result $\hat{y}=\text{FC}_\text{cls}(\mE_\text{agg})$. The expression classification loss $\gL_\text{cls}$ is defined as the cross-entropy loss between the prediction result $\hat{y}$ and the ground truth $y$.

\vspace{-0.2cm}
\subsection{Loss Function}
In our method, the spectrum classification loss $\gL_\text{spectrum}$ from the SAOD module and the expression classification loss $\gL_\text{cls}$ from the HGFE module are jointly optimized in an end-to-end manner. The overall loss function is defined as
\begin{equation}
     \gL = \gL_\text{cls} + \lambda\gL_\text{spectrum}
\end{equation}
where $\lambda$ is the balance parameter between the two parts. Under the supervision of the joint loss function, the model learns the modality-invariant and discriminative feature representation simultaneously for better NIR-VIS facial expression recognition.
\vspace{-0.2cm}

\section{Experiment Setup}
In this section, we present the details of datasets, task settings, baselines and implementation.

\vspace{-0.2cm}
\subsection{Datasets}
We conduct experiments on two NIR\&VIS facial expression datasets: Oulu-CASIA \cite{zhao2011facial} and the proposed Large-HFE. 
There are rather few publicly available datasets in the field of NIR-VIS heterogeneous FER \cite{yu2022cmos}.
To the best of our knowledge, \textbf{Oulu-CASIA} is the only publicly available dataset before this paper, which consists of NIR\&VIS video sequences captured in the lab environment with 6 basic facial expressions and 80 subjects. Among the subjects, there are 21 females and 59 males. Following \cite{ruan2021feature}, we select the last 3 frames in each sequence of strong illumination to construct the image dataset. 

\textbf{Large-HFE} is a new NIR\&VIS facial expression dataset proposed in this paper. Compared with Oulu-CASIA, the proposed dataset has larger size in subjects, wider range in age distribution, and more diversity in cultural background. In Large-HFE, the images are also captured in the laboratory-controlled environment, with 6 basic expressions plus neutral and 360 subjects at the age of 5-70. Among the subjects, there are 184 females and 176 males. The subjects are from different cultural backgrounds, including Asia, Europe, Africa and Americas. For each expression of each subject, 1-2 images captured in the normal illumination are selected to construct the dataset. 

To enhance the training set with extra VIS samples, we select 2 VIS facial expression datasets: KDEF \cite{lundqvist1998karolinska} and CFEE \cite{du2014compound}. \textbf{KDEF} accommodates 70 subjects with 6 basic expressions plus neutral, in which images from the front view are utilized in the experiments. \textbf{CFEE} consists of 230 subjects with basic and compound expressions. For comparability with the NIR\&VIS datasets, the selected datasets are both from the laboratory-controlled environment and only samples of 6 basic expressions are selected.
\vspace{-0.2cm}
\subsection{Baselines and Task Settings}
To verify the effectiveness of the proposed method, we select baselines from two related tasks: single-modality FER methods and heterogeneous face recognition methods. For the single-modality FER methods, we select four baselines including IL-CNN \cite{cai2018island}, SCN \cite{wang2020suppressing}, MA-Net \cite{zhao2021learning}, and FDRL \cite{ruan2021feature}. For the heterogeneous face recognition (HFR) methods, we select two subspace learning methods including DSVNs \cite{hu2019disentangled} and DFD \cite{hu2022domain}, and two image synthesis methods including HiFaceGAN \cite{yang2020hifacegan} and T2V-DDPM \cite{nair2023t2v}. 

In the experiments, the training set of each NIR\&VIS facial expression dataset (original set) and the full set of the above 2 VIS facial expression datasets (extra set) are mixed to form the enhanced training set (enhanced set).
For the train set of \textbf{NIR only}, only the NIR samples in the original set are utilized to train the single-modality FER models. For the train set of \textbf{NIR + VIS}, both NIR and VIS samples from the original set is utilized to train the single-modality FER models and subspace learning HFR models. For the train set of \textbf{NIR + VIS + Extra VIS}, both NIR and VIS samples from the enhanced set are utilized to train all the models. 
The test set of NIR\&VIS facial expression dataset is used as the test set for all settings. The classification accuracy and macro F1 score are evaluated on the NIR part of the test set.

\begin{table*}[t]
\vspace{-5mm}
    \centering
    \caption{The performance on two NIR\&VIS facial expression datasets in $\%$. \textit{NIR only} means only using the NIR part of the dataset as the train set. \textit{NIR + VIS} means using both NIR and VIS parts of the dataset as the train set. \textit{NIR + VIS + Extra VIS} means using the dataset and extra samples from VIS facial expression datasets as the train set. \dag\ denotes single-modality FER methods. \ddag\ denotes heterogeneous face recognition (HFR) methods. The best results are highlighted in \textbf{bold}.}
\vspace{-2mm}
\begin{tabular}{c|l|cc|cc}
    \toprule
    \multirow{2}[1]{*}{Train Set} & \multicolumn{1}{c|}{\multirow{2}[1]{*}{Method}} & \multicolumn{2}{c|}{Oulu-CASIA} & \multicolumn{2}{c}{Large-HFE} \\
          &       & Acc   & F1    & Acc   & F1 \\
    \midrule
    \multirow{5}[2]{*}{NIR only} & ResNet \cite{he2016deep} \dag & $76.39_{\pm0.55}$ & $76.32_{\pm0.71}$ & $74.23_{\pm0.67}$ & $74.07_{\pm1.12}$ \\
          & IL-CNN \cite{cai2018island} \dag & $72.71_{\pm0.21}$ & $72.36_{\pm0.52}$ & $72.08_{\pm0.44}$ & $71.94_{\pm0.69}$ \\
          & FDRL \cite{ruan2021feature} \dag & $70.29_{\pm0.38}$ & $69.71_{\pm0.48}$ & $71.95_{\pm1.55}$ & $70.75_{\pm1.40}$ \\
          & MA-Net \cite{zhao2021learning} \dag & $71.46_{\pm1.73}$ & $71.40_{\pm1.16}$ & $73.96_{\pm0.38}$ & $73.70_{\pm0.60}$ \\
          & SCN \cite{wang2020suppressing} \dag & $78.86_{\pm0.69}$ & $78.78_{\pm0.69}$ & $79.34_{\pm0.62}$ & $79.23_{\pm0.62}$ \\
    \midrule
    \multirow{8}[2]{*}{NIR + VIS} & ResNet \cite{he2016deep} \dag & $78.75_{\pm0.76}$ & $78.91_{\pm0.83}$ & $77.28_{\pm0.91}$ & $77.30_{\pm0.90}$ \\
          & IL-CNN \cite{cai2018island} \dag & $76.25_{\pm0.49}$ & $76.23_{\pm0.59}$ & $76.18_{\pm0.34}$ & $76.35_{\pm0.70}$ \\
          & FDRL \cite{ruan2021feature} \dag & $76.67_{\pm0.31}$ & $76.63_{\pm0.49}$ & $77.15_{\pm1.71}$ & $77.49_{\pm1.59}$ \\
          & MA-Net \cite{zhao2021learning} \dag & $68.68_{\pm1.47}$ & $68.74_{\pm1.40}$ & $70.40_{\pm0.53}$ & $70.34_{\pm0.57}$ \\
          & SCN \cite{wang2020suppressing} \dag & $80.04_{\pm0.95}$ & $79.97_{\pm0.51}$ & $80.04_{\pm0.68}$ & $79.79_{\pm0.33}$ \\
          & DSVNs \cite{hu2019disentangled} \ddag & $72.57_{\pm1.14}$ & $73.08_{\pm1.12}$ & $71.01_{\pm0.72}$ & $70.07_{\pm0.79}$ \\
          & DFD \cite{hu2022domain} \ddag & $64.58_{\pm1.74}$ & $63.94_{\pm1.78}$ & $77.78_{\pm1.32}$ & $77.68_{\pm1.36}$ \\
          & NFER-Former (proposed) & $\bm{81.95}_{\pm0.60}$ & $\bm{81.86}_{\pm0.65}$ & $\bm{80.49}_{\pm0.90}$ & $\bm{80.11}_{\pm1.00}$ \\
    \midrule
    \multirow{10}[1]{*}{NIR + VIS + VIS Extra} & ResNet \cite{he2016deep} \dag & $77.64_{\pm0.78}$ & $77.83_{\pm0.90}$ & $76.43_{\pm0.91}$ & $76.41_{\pm0.91}$ \\
          & IL-CNN \cite{cai2018island} \dag & $79.17_{\pm0.68}$ & $79.15_{\pm0.75}$ & $76.59_{\pm0.63}$ & $76.74_{\pm0.72}$ \\
          & FDRL \cite{ruan2021feature} \dag & $78.82_{\pm0.61}$ & $78.69_{\pm0.61}$ & $79.12_{\pm1.45}$ & $78.92_{\pm1.55}$ \\
          & MA-Net \cite{zhao2021learning} \dag & $73.54_{\pm1.40}$ & $73.70_{\pm1.42}$ & $72.67_{\pm0.36}$ & $72.48_{\pm0.36}$ \\
          & SCN \cite{wang2020suppressing} \dag & $80.53_{\pm0.71}$ & $80.61_{\pm0.72}$ & $79.48_{\pm0.55}$ & $79.23_{\pm0.54}$ \\
          & DSVNs \cite{hu2019disentangled} \ddag & $76.39_{\pm1.07}$ & $61.56_{\pm1.06}$ & $72.81_{\pm0.52}$ & $72.32_{\pm0.51}$ \\
          & DFD \cite{hu2022domain} \ddag & $66.67_{\pm1.43}$ & $66.21_{\pm1.48}$ & $79.83_{\pm1.04}$ & $79.82_{\pm1.13}$ \\
          & HiFaceGAN \cite{yang2020hifacegan} \ddag & $75.83_{\pm1.21}$ & $75.76_{\pm1.19}$ & $77.31_{\pm1.20}$ & $77.36_{\pm1.27}$ \\
          & T2V-DDPM \cite{nair2023t2v} \ddag & $73.15_{\pm1.30}$ & $73.02_{\pm1.30}$ & $74.80_{\pm1.44}$ & $74.70_{\pm1.04}$ \\
          & NFER-Former (proposed) & $\bm{84.03}_{\pm0.65}$ & $\bm{83.82}_{\pm0.71}$ & $\bm{82.17}_{\pm1.08}$ & $\bm{81.99}_{\pm1.18}$ \\
    \bottomrule
    \end{tabular}%
    
    \label{tab:main}%
    \vspace{-4mm}
\end{table*}%

\vspace{-0.2cm}
\subsection{Implementation Details}
The proposed method is implemented with \textit{PyTorch} 1.12. The backbone is Vision Transformer \cite{dosovitskiy2020image}. Before training, the facial areas of the image samples are detected and cropped by the face detector in \textit{dlib} library. The cropped images are then resized and randomly cropped to $224 \times 224$ and the batch size is set to 64. The training process is stopped after 40 epochs. The AdamW optimizer is adopted to optimize the backbone with the learning rate set to $10^{-4}$ and a one-cycle scheduler. The weight decay is set to $5\times10^{-4}$. The hyper-parameter $\lambda$ is set to $0.1$ for both Oulu-CASIA and Large-HFE. For each experiment, subject-independent five-fold cross-validation protocol is adopted and the average results are reported in this paper.

\vspace{-0.2cm}
\section{Results and Analysis}
In this section, we present the experimental results and analysis. First, we compare the proposed method with the baselines on the two NIR\&VIS facial expression datasets. Then, we conduct ablation study to verify the effectiveness of each component in the proposed method. 
\vspace{-0.2cm}
\subsection{Comparison with State-of-the-art Methods}

Table \ref{tab:main} shows the performance of the proposed method and several state-of-the-art baselines on Oulu-CASIA and Large-HFE under different settings of train set. The compared baselines are categorized into two groups, including facial expression recognition baselines and heterogeneous face recognition baselines. From the results, we can obtain the following observations.

(1) It is significant that the proposed framework achieves the best performance on both datasets with the settings of NIR + VIS and NIR + VIS + Extra VIS. Compared with the best baseline with NIR + VIS + Extra VIS setting, the proposed method improves the accuracy by $3.50\%$ and $2.34\%$, and the macro F1 score by $3.21\%$ and $2.17\%$ on Oulu-CASIA and Large-HFE respectively. The results demonstrate the effectiveness of the proposed method for NIR FER with the augmentation of visible facial expression datasets. This is mainly due to the proposed SAOD module and HGFE module, which can perform effective disentanglement of spectrum information using vision transformer and extract the discriminative feature embeddings with the guidance of knowledge-hypergraph. Therefore, the proposed method can make full use of the visible samples to enhance the performance of NIR FER. 

(2) Among the compared baselines, FER methods tend to show higher performance than HFR methods. FER baselines including SCN, IL-CNN and FDRL achieve relatively high performance on both datasets, since they are specially designed to address the challenges in FER task, such as inter-class similarity and noisy labels. The performance of HFR methods such as DSVNs and T2V-DDPM is relatively less satisfactory in most situations since the consideration for subtle facial representations and high-order inter-class correlations is lacking for facial expression modeling.

\vspace{-0.2cm}
\subsection{Ablation Study}
\begin{table}
    \centering
    \caption{The ablation study results on Oulu-CASIA and Large-HFE in $\%$. The best results are highlighted in \textbf{bold}.}
    \vspace{-2mm}
    \setlength{\tabcolsep}{2.5mm}{
    \begin{tabular}{l|cccc}
        \toprule
        \multicolumn{1}{c|}{\multirow{2}[1]{*}{Method}} & \multicolumn{2}{c}{Oulu-CASIA} & \multicolumn{2}{c}{Large-HFE} \\
            & Acc & F1 & Acc & F1 \\
        \midrule
        Baseline & $81.21$ & $81.22$ & $79.11$ & $78.89$ \\
        SAOD  & $82.71$ & $82.75$ & $80.51$ & $80.20$ \\
        HGFE  & $82.15$ & $82.03$ & $80.78$ & $80.59$ \\
        SAOD + HGFE & $\bm{84.03}$ & $\bm{83.82}$ & $\bm{82.17}$ & $\bm{81.99}$ \\
        \bottomrule    
    \end{tabular}%
    }
    \label{tab:ablation-acc}%
    \vspace{-0.3cm}
\end{table}%

In order to verify the effectiveness of each component in the proposed method, ablation study is conducted on Oulu-CASIA and Large-HFE with different combinations of the key modules.
Specifically, 4 combinations of key modules are compared in the ablation study as the followings:
\begin{itemize}
    \item \textbf{Baseline}: The basic ViT backbone.
    \item \textbf{SAOD}: The basic model with SAOD module.
    \item \textbf{HGFE}: The basic model with HGFE module.
    \item \textbf{SAOD + HGFE}: The basic model with both SAOD and HGFE modules, i.e., the proposed method.
\end{itemize}

As shown in Table \ref{tab:ablation-acc}, the ViT backbone performs the worst, which is limited by the modality gap and subtle inter-class variation in NIR FER. 
Incorporating the SAOD module into ViT backbone significantly improves the performance, which indicates that the SAOD module effectively disentangles the spectrum information from the modality-invariant feature extraction. Moreover, the performance is further improved by incorporating the HGFE module, which demonstrates the effectiveness of the high-order inter-class correlation from knowledge hypergraph. The proposed framework that combines SAOD and HGFE modules makes full use of the above advantages and achieves the best performance.

\begin{figure}[htbp]
    \centering
        \vspace{-2mm}
    \begin{tikzpicture}[scale=0.8]
        \begin{axis}[
            ymax=90,
            ymin=75,
            xlabel=$\lambda$,
            ylabel=Accuracy( $\%$ ),
            xtick={-2,-1,0,1,2},
            xticklabels={0.01,0.1,1,5,10},
            tick align=inside,
            legend style={draw=none, fill=none},
            y tick label style={
            /pgf/number format/.cd,
            fixed,
            fixed zerofill,
            precision=2,
        },
        ]
        \addplot[ mark=triangle*,mark options={scale=2, draw=red, fill=red, rotate=180,  fill opacity=0},red]plot coordinates{
            (-2,82.29)
            (-1,84.028)
            (0,83.194)
            (1,83.4)
            (2,80.092)
            };
        \node [above] at (-1,84.028) {84.03$\%$};
        \addlegendentry{Oulu-CASIA}

        \addplot[ mark=square,blue,mark options={scale=1} ]plot coordinates{
            (-2,80.356)
            (-1,82.174)
            (0,80.628)
            (1,80.93)
            (2,79.692)
            };
        \node [above] at (-1,82.174) {82.17$\%$};
        \addlegendentry{Large-HFE}
        \end{axis}
    \end{tikzpicture}
    \caption{Ablation study results for different values of hyper-parameter $\lambda$ on Oulu-CASIA and Large-HFE.}
    \vspace{-5mm}
    \label{fig:lambda}
\end{figure}
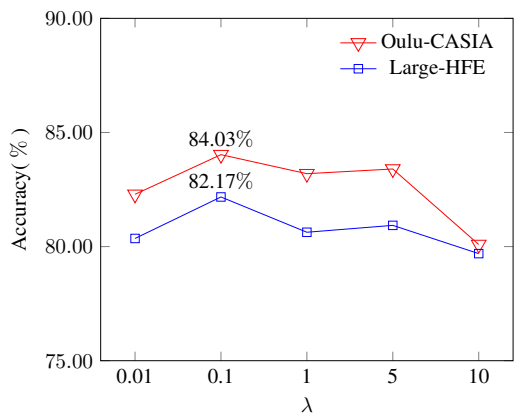

In order to verify the recognition performance of the proposed method with different values of hyper-parameter $\lambda$, we conduct ablation study of the hyper-parameter $\lambda=0.01, 0.1, 1, 5, 10$ on Oulu-CASIA and Large-HFE, as shown in Fig. \ref{fig:lambda}. It is observed that our method achieves the best performance when $\lambda=0.1$ for both datasets. Therefore, we set the values of $\lambda$ to $0.1$ in the experiments.

\vspace{-0.2cm}
\subsection{Visualization}

\begin{figure}[t]
\vspace{-0.2cm}
    \centering
    \includegraphics[width=0.85\columnwidth]{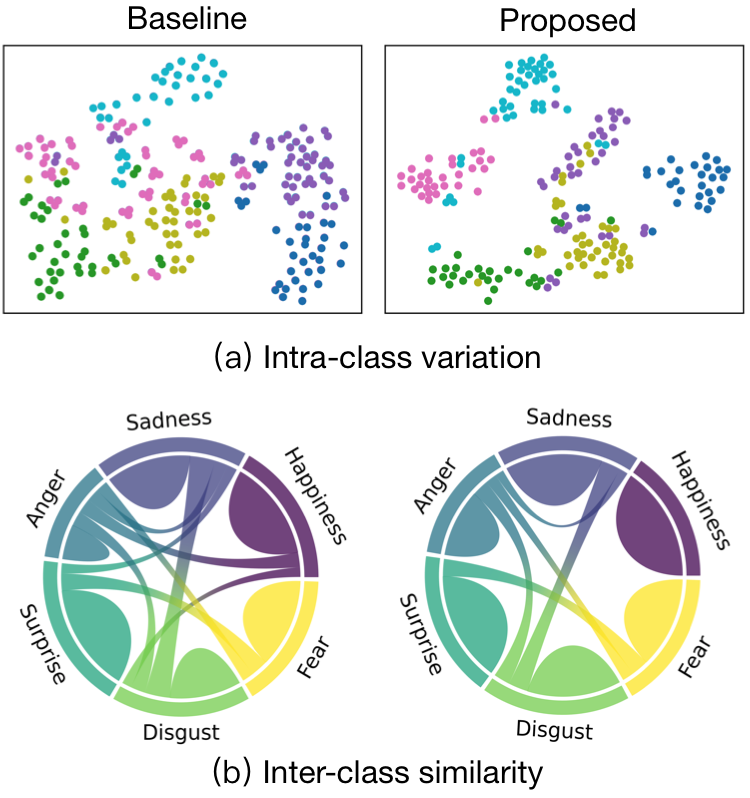} 
    \caption{Visualization of intra-class variation and inter-class similarity of the expression features learned by baseline and proposed method. Different colors denote different classes of expressions.}
    \label{fig:feature-scatter}
    \vspace{-0.3cm}
\end{figure}

In order to demonstrate the performance on the challenges of intra-class variation and inter-class similarity, we visualize the expression features extracted by the baseline and the proposed NFER-Former on Large-HFE dataset in Fig. \ref{fig:feature-scatter}.

As shown in Fig. \ref{fig:feature-scatter}.a, the intra-class variations of the extracted features are illustrated in the scatter plot, where the features are reduced to 2D space using t-SNE \cite{van2008visualizing}.
From the figure, we can see that the features learned by the baseline are not compact for each expression class. In contrast, the features learned by the proposed method show more compact intra-class variation.

As shown in Fig. \ref{fig:feature-scatter}.b, the inter-class similarities of the extracted features are illustrated in the chord diagram, where each expression class is regarded as a node and the feature similarity is regarded as the edge weight between two classes. The similarity between two classes is calculated from the average distance from the features of one class to the features of another class. For clarity, the edges with similarity score less than a constant threshold are removed.
It is observed that features extracted by baseline method have higher inter-class similarity, such as \textit{happiness} and \textit{anger}, \textit{sadness} and \textit{surprise}. In contrast, the proposed method effectively alleviates the inter-class similarity problem.

\vspace{-0.2cm}
\section{Conclusion}
In this paper, we propose a novel NFER-Former method to utilize NIR-VIS heterogeneous facial expression samples for effective NIR FER. We introduce Self-Attention Orthogonal Decomposition and Hypergraph-Guided Feature Embedding modules to learn the modality-invariant discriminative features from NIR-VIS heterogeneous samples. We construct a new dataset named Large-HFE with 4.5 times more subjects than the previous dataset for better evaluation of NIR FER. The experimental results show that our NFER-Former method achieves state-of-the-art performance on existing NIR FER datasets.

\bibliography{aaai24}

\end{document}